\RequirePackage{fix-cm}
\documentclass[smallextended]{svjour3}       
\smartqed  
\usepackage{graphicx}
\usepackage{graphicx}
\usepackage{amssymb}
\usepackage{subfigure}
\usepackage{graphics}
\usepackage[normalem]{ulem}
\usepackage{url}
\usepackage{epsfig}
\usepackage{algorithm}
\usepackage{pstricks}
\usepackage{xspace}
\usepackage{float}
\usepackage{enumitem}
\usepackage{latexsym}
\usepackage{color}
\usepackage{colortbl}
\usepackage{epstopdf}
\usepackage{caption}
\usepackage{multirow}
\usepackage{placeins}
\usepackage{dsfont}
\usepackage{hhline}
\usepackage{balance}
\usepackage{enumitem}
\usepackage[euler]{textgreek}
\usepackage{caption}
\usepackage[noend]{algpseudocode}
\usepackage{mathrsfs}
\usepackage{tabularx}
\usepackage{mathtools}
\usepackage{ulem}
\usepackage{booktabs}

\begin{document}

\title{Binarized Graph Neural Network
}


\author{Hanchen Wang         \and
        Defu Lian \and 
        Ying Zhang \and 
        Lu Qin \and
        Xiangjian He \and
        Yiguang Lin \and
        Xuemin Lin
}


\institute{Hanchen Wang \at
             CAI, University of Technology, Sydney, Australia \\
              \email{hanchenw.au@gmail.com}           
           \and
           Defu Lian \at
           University of Science and Technology of China \\
           \email{liandefu@ustc.edu.cn}
           \and
           Ying Zhang \at
           CAI, University of Technology, Sydney, Australia \\
            \email{Ying.Zhang@uts.edu.au}
            \and
            Lu Qin \at
            CAI, University of Technology, Sydney, Australia \\
              \email{Lu.Qin@uts.edu.au}
              \and
            Xiangjian He \at
            University of Technology, Sydney, Australia \\
              \email{Xiangjian.He@uts.edu.au}
              \and
           Yiguang Lin \at
           University of Technology, Sydney, Australia \\
              \email{Yiguang.Lin@uts.edu.au}
              \and
           Xuemin Lin \at
           University of New South Wales \\
           \email{lxue@cse.unsw.edu.au}
}

\date{Received: date / Accepted: date}

\maketitle

\begin{abstract}
Recently, there have been some breakthroughs in graph analysis by applying the graph neural networks (GNNs) 
following a neighborhood aggregation scheme, which demonstrate outstanding performance in many  tasks.
However, we observe that the parameters of the network and the embedding of nodes are represented in real-valued matrices 
in existing GNN-based graph embedding approaches which may limit the efficiency and scalability of these models.
It is well-known that binary vector is usually much more space and time efficient than the real-valued vector. 
This motivates us to develop a binarized graph neural network to learn the binary representations of the nodes 
with binary network parameters following the GNN-based paradigm.
Our proposed method can be seamlessly integrated into the existing GNN-based embedding approaches to binarize the model parameters and learn the compact embedding. 
Extensive experiments indicate that the proposed binarized graph neural network, namely BGN, is orders of magnitude more efficient in terms of both time and space while matching the state-of-the-art performance.
\keywords{Graph Neural Network \and Binarized Neural Network \and Classification}
\end{abstract}

\section{Introduction}
\label{sec:intro}

Graph analysis provides powerful insights into how to unlock the value graphs hold. 
Due to this power, techniques
for analyzing graphs are becoming an increasingly popular topic of study in both academics and industry.
To effectively and efficiently support important analytic tasks on graph data, such as node/graph classification, node clustering, community detection, node recommendation, link prediction and graph visualization, a variety of graph embedding techniques
(See~\cite{DBLP:journals/debu/HamiltonYL17,DBLP:journals/tkde/CuiWPZ19} for a comprehensive survey) have been developed.
Graph data is mapped into low-dimension data such that the proximity relationship among graph nodes (i.e., objects) is preserved and the off-the-shelf machine learning methods, which are designed to handle vector representations, can be immediately applied.

The existing graph embedding techniques can be roughly classified into three broad categories:
(1) random walk based embedding (e.g., Deepwalk~\cite{perozzi2014deepwalk} and Node2vec~\cite{DBLP:conf/kdd/GroverL16}) ;
(2) node similarity based embedding (e.g., LINE~\cite{DBLP:conf/www/TangQWZYM15}
and NetMF~\cite{DBLP:conf/wsdm/QiuDMLWT18}); and (3) graph neural networks (GNN) based embedding (e.g., GCN~\cite{kipf2017semi}, GraphSage~\cite{hamilton2017inductive}, GAT~\cite{velivckovic2017graph} and AS-GCN~\cite{DBLP:conf/nips/Huang0RH18}).
As reported by Leskovec et al. in their tutorial on graph embedding at WWW 2018\footnote{\url{http://snap.stanford.edu/proj/embeddings-www}}, the first two categories of embedding techniques are only able to learn a ``shallow'' representation of the graph nodes due to the simplicity of the models.
It is shown in~\cite{kipf2017semi,hamilton2017inductive} that the neural network based embedding methods significantly outperform the state-of-the-art techniques in the first two categories for the node classification task.
Therefore, exploring how to use neural network to create a ``deep'' representation more efficiently is a promising direction in graph representation learning.
However, most of the existing graph neural network models suffer from the scalability issue due to the high time and space cost of the real-valued model.

Recently, there have been some researches on learning binary graph embedding (e.g.,~\cite{DBLP:conf/kdd/Lian0ZGCT018,DBLP:conf/ijcai/PLOS18,DBLP:conf/icdm/YangPLZ18}),
in which each node is represented by a binary vector (code), instead of a real-valued vector.
It has been shown that the binarized graph embedding can achieve much better time and space efficiency.

\vspace{1mm}
\noindent {\bf Time efficiency}. It is well-known that the distance computation of binary vectors
(i.e., Hamming distance) is much more efficient than that of real-valued vectors (e.g., Euclidian distance).
In addition to the specifically tailored search algorithms (e.g.,~\cite{DBLP:conf/icde/QinWXWLI18}),
the dot product between binary vectors can also enjoy the hardware support
(e.g., \textit{xnor} and build-in CPU instruction \textit{popcount}).

As stressed in a recent work~\cite{DBLP:conf/icml/LiGDVK19} from \textit{DeepMind}, the pairwise dot product of the vectors has been intensively used by the model for some specific tasks (e.g., graph similarity computation in~\cite{DBLP:conf/wsdm/BaiDBCSW19}).
Thus, the binary vector has been used in their graph matching network (GMN) to speedup the computation.

\vspace{1mm}
\noindent {\bf Space Efficiency}.
The binary embedding can represent the node in a compact way while well preserving the structure information. As shown in~\cite{DBLP:conf/kdd/Lian0ZGCT018}, INH-MF can achieve competitive graph node classification performance with 128 bits for each node compared to the conventional embedding approaches
(e.g., DeepWalk) with 128 dimensions (i.e., $128 \times 64$ bits) per node.
This will be a great advantage when we face a large-scale graph because the binarized embedding of a graph is more likely to be accommodated in the main memory.



\vspace{1mm}
\noindent \textbf{Motivation and Challenges.}
The existing GNN-based methods have demonstrated outstanding performance in various tasks
such as classification~\cite{hamilton2017inductive,kipf2017semi,velivckovic2017graph,DBLP:conf/nips/Huang0RH18}, link prediction~\cite{DBLP:conf/nips/ZhangC18,DBLP:conf/nips/Kazemi018}, graph similarity match~\cite{DBLP:conf/wsdm/BaiDBCSW19,DBLP:conf/icml/LiGDVK19} and graph clustering~\cite{DBLP:conf/ijcai/WangPHLJZ19,DBLP:conf/ijcai/0003LLW19}.
However, they may suffer from the limitation of the memory and speed due to the use of real-valued vectors
for node and graph representations and model parameters.


Given the outstanding embedding quality, various applications of the GNN-based approaches and the space and time efficiency of the binarized representation, one may wonder if we can design a binarized GNN-based graph embedding approach
such that we can achieve a good trade-off between embedding quality and time/space efficiency in the GNN-based methods.

We notice that the existing binarized graph embedding methods~\cite{DBLP:conf/kdd/Lian0ZGCT018,DBLP:conf/ijcai/PLOS18} rely on the discretization of the matrix factorization following the node-similarity based approaches.
They cannot be extended to binarize the GNN-based embedding due to the inherently different natures of two categories of approaches.

As to our best knowledge, the only attempt for the binarization of GNN is from \textit{DeepMind} in their recent work~\cite{DBLP:conf/icml/LiGDVK19}. Their binarization method converts each learned $d$-dimensional real-valued vector
into a $d$-dimensional "nearly" binary vector by applying well-known binarization function \textit{tanh}
to approximate hamming distance for the binarization and optimization.
However, the output of \textit{tanh} is not exact binary value and cannot be accelerated by the binary logic operations
(e.g., xnor and popcount).
As an alternative, one may consider the Binarized Neural Network (BNN) (e.g.,~\cite{DBLP:conf/nips/HubaraCSEB16}) for the graph embedding so that the representation is naturally binarized.
However, BNN is not designed for graph data, and as to our best knowledge, there is no existing graph embedding work
based on BNN.



These issues motivate us to develop a new binarized graph embedding technique which can be integrated into existing GNN-based models to binarize the parameters and produce high-quality binarized graph embeddings.
The key challenge is how to generate effective compact embedding vectors with binary network parameters in an effective way. 
To address the challenge, we design a binarized graph neural network framework to learn the binary parameters and representations efficiently and effectively .

\noindent \textbf{Contributions.}
Our principle contributions are summarized as follows:
\vspace{-2mm}
\begin{itemize}

\item To the best of our knowledge, this is the first study on binarized graph neural network (GNN) with binary parameters to generate binary graph representations. The proposed method, namely \textbf{BGN}, can be seamlessly integrated into the existing GNNs.

\item An end-to-end binarized graph neural network framework is proposed with binary weights and activations.
This binarized framework can immediately reduce the memory consumption for the network; 
the bit-wise operations between the binary vectors can substantially speedup the inference time of the model
and the gradient estimator enables our model to effectively process back-propagation through discrete parameters and activations.


\item Extensive experiments on multiple benchmark networks are conducted for node classification task. The results demonstrate that our proposed method outperforms existing binarized embedding methods with a big margin.
Compare to the real-valued GNNs, our BGN model can achieve nearly state-of-the-art performance while consuming much fewer computation resources (up to $1/28$ parameter and embedding memory space and $1/20$ inference time).

\item Binarization approaches are employed on the GNN-based application GMN to show that,
      by applying our BGN techniques, GMN model can dramatically reduce the time and space complexity 
      while keeping the performance competitiveness.
   
\item Experiments further show that our proposed BGN technique allows users to achieve a trade-off 
between the space/time and embedding quality in a flexible way by tuning different level and setting of binarization 
on the parameters and activations.

\end{itemize} 
\section{Related Works}
\label{sec:related}

\vspace{1mm}
\noindent \textbf{Graph Embedding.}
A key problem in machine learning on graphs is finding a way to incorporate information about the structure of the graph into the desired machine learning model.
Graph embedding is one of the most promising approaches because it maps nodes into a low-dimensional space such that the structure of the graph is well preserved.
Once accomplished, an existing machine learning approach (e.g., k-means clustering) can be used to assimilate and analyse the graph in the embedded low-dimensional space. Loosely following the seminal graph embedding approach, DeepWalk,
three broad categories of embedding methods have appeared in the literature:
(1) node similarity based embedding methods (e.g., LINE, NetMF),
which rely on the proximity of the nodes w.r.t various similarity metrics.
The matrix factorization techniques have been used to learn the embedding of the nodes.
(2) Random walk based embedding methods (e.g., Deepwalk
and node2vec) which encode the nodes by applying the Skip-Gram technique~\cite{mikolov2013distributed} on the random walks;
and
(3) graph neural networks (GNN) based embedding methods
(e.g., GCN, GraphSage and GIN)
which apply the neural network techniques on graph to learn the representations of the nodes.

Most of the existing graph embedding studies use the real-valued vector to encode the graph nodes
following the above three computing paradigms.
Recently, three unsupervised approaches~\cite{DBLP:conf/kdd/Lian0ZGCT018,DBLP:conf/ijcai/PLOS18,DBLP:conf/icdm/YangPLZ18} have been proposed to learn the \textit{binary embedding} of the graphs following the node-similarity based embedding methods.
Particularly, INH-MF~\cite{DBLP:conf/kdd/Lian0ZGCT018} and
DNE~\cite{DBLP:conf/ijcai/PLOS18} are independently developed for binarized graph embedding based on the discretization of the matrix factorization on proximity graphs.
BANE proposed in~\cite{DBLP:conf/icdm/YangPLZ18} is a natural extension of DNE by considering both structure and attribute similarities on the attributed graphs.

\vspace{1mm}
\noindent \textbf{Binary Hashing.}
The binary hashing has been widely used to learn the binary vectors (codes) of the objects in many applications.
The most popular application is the approximate nearest neighbor search in high dimension space where binary hashing methods encode high-dimensional objects (e.g., documents and images)
to binary codes, while preserving similarity distance in the original space.
Many learning to hash approaches have been proposed including unsupervised methods
(e.g., ~\cite{DBLP:journals/ijar/SalakhutdinovH09,DBLP:conf/nips/LiuMKC14}),
supervised methods (e.g.,~\cite{DBLP:conf/cvpr/ShenSLS15}),
and deep learning based methods (e.g.,~\cite{DBLP:conf/cvpr/Liu0SC16}).
Please refer to~\cite{DBLP:journals/pami/WangZSSS18} for a comprehensive survey.
Recently, three approaches~\cite{DBLP:conf/kdd/Lian0ZGCT018,DBLP:conf/ijcai/PLOS18,DBLP:conf/icdm/YangPLZ18} have been proposed to learn the binary embedding of the graphs following the node-similarity based embedding methods.
As to our best knowledge, there is no existing work on the binarized graph embedding
based on GNNs.

\vspace{1mm}
\noindent \textbf{Binarized Neural Networks}
Binarized neural networks was first proposed by BNN~\cite{courbariaux2016binarized}.
The binarization technique proposed in \cite{courbariaux2016binarized} is used by most network binarization models.
Among them, XNOR-Net~\cite{rastegari2016xnor} and DoReFa-Net~\cite{DBLP:journals/corr/ZhouNZWWZ16} are the most popular ones because of their great performance on the image classification task.

XNOR-Net was proposed to have high accuracy of classification task on the ImageNet dataset while XNOR-Net has $58\times$ faster convolutional operations and $32\times$ memory saving.
DoReFa-Net replaces the binarization by quantization which allows the model to change the bit size for weights, activations and even gradient calculations during backpropagation.

However, these methods are all designed for computer vision tasks.
Though they perform well on the image dataset, they cannot be adapted to the graph representation learning and graph analysis task directly.

\vspace{1mm}
\noindent \textbf{Graph Neural Network Applications}
There are several applications that are based on the GNN.
Such as Graph Matching Network~\cite{DBLP:conf/icml/LiGDVK19} and SimGNN~\cite{DBLP:conf/wsdm/BaiDBCSW19}.
These models utilize GNN and use the similarity (distance) of graph embedding to approximate the graph edit distance and graph similarity.

The Graph Matching Network (i.e., GMN) is a novel GNN-based framework proposed by \textit{DeepMind} to compute the similarity score between input pairs of graphs.
Separate MLPs will first map the input nodes in the graphs into vector space.
Then the propagation layer will aggregate the messages of the edges and cross-graph matching vector by MLP or GRU with input concatenation of node representations and edge vectors.
Matching function is applied to compute the attention coefficients based on the node information between the input pair of graphs.
The matching function is based on the softmax function over node vectors which requires the calculation of vector space similarity like Euclidean, cosine similarity or dot product between all pairs of node representations.
This attention coefficients calculation across two graphs requires a computation cost of $\mathcal{O}(\mid V_1 \mid \mid V_2 \mid d)$, where $V_1$ and $V_2$ indicate the number of vertices of input graph 1 and 2 respectively, and $d$ is the dimension of the node representation.
The match vector ${\bf \mu}_{j\rightarrow i}$ is concatenated with the message vector ${\bf m}_{j\rightarrow i}$ and the node representation ${\bf h}^{(t)}_{i}$, then the concatenation is fed into MLP or a recurrent neural network core to produce the new node representations.
Given the learned node representations of graph, the aggregation module proposed in~\cite{li2015gated} is used to obtain the graph representations.
The similarity score in vector space such as Euclidean similarity, cosine similarity and approximate hamming similarity will be computed   between graph representations to approximate the similarity between the input graphs.

\section{Background and Preliminaries}
\label{sec:pre}

Recent studies have revealed that graph neural network can perform excellently on label classification tasks.
The existing GNN-based graph embedding approaches share the same computing paradigm.
GNNs take graph nodes' feature and neighborhood information as the input.
During the training, the representations of nodes (real-valued vectors) at each layer will be updated
by the aggregators and non-linear activation functions. 
The output representations will be fed into the task-specific layer to calculate the loss of the model.
Based on that, the model will be optimized by the optimizer through backpropagation.
The main differences among these GNN-based graph embedding approaches are the design of the aggregator which combines the context representations and the loss function designed for different graph analytic tasks.

These models have real-valued parameters and learn a real-valued representation for each node in an end-to-end manner for graph node classification.
However, the real-valued parameters and representations are space-consuming for storage and time-consuming for multiplication computation, especially for large-scale graphs. 
To address these issues, in this paper we devise a novel binarized graph neural network, namely BGN, with binary parameters in the neural network to learn binary embedding representations for node classification task.

The important notations used throughout the paper are summarized in Table~\ref{tab:notations}.

\begin{table}
\small
  \centering
    \begin{tabular}{|p{0.15\columnwidth}|p{0.7\columnwidth}|}
      \hline
      \textbf{Notation}   & \textbf{Definition}                   \\ \hline \hline

      $G$    &  the graph dataset  \\  \hline
      $\mathcal{V}, E$    &  the set for nodes and edges in the graph.  \\ \hline
     ${\bf x}_v$    &  the feature information for node $v$.  \\ \hline
     ${\bf \eta}_v$    &  the neighborhood nodes of node $v$.  \\ \hline
    ${(\cdot)}^{\bf b}$    &  denotes that the vector or matrix is binary-valued.  \\ \hline
    ${\bf h}_v$    &  the hidden representation of node $v$.  \\ \hline
    ${\bf W}$    &  the weight matrix in the neural network.  \\ \hline
    $\mathcal{B}(\cdot)$    &  the binarization function which is used to transform the real-valued vector or matrix into binary-valued vector or matrix.  \\ \hline
    $\alpha_{ij}$    &  the attention coefficient between node $i$ and node $j$.  \\ \hline
\end{tabular}
\vspace{2mm}
 \caption{Summary of Notations}
\vspace{-8mm}
 \label{tab:notations}
\end{table}
\section{Binarized Graph Neural Network}

\begin{figure*}[tb]
\vspace{-2mm}
\centering
\includegraphics[width=\columnwidth]{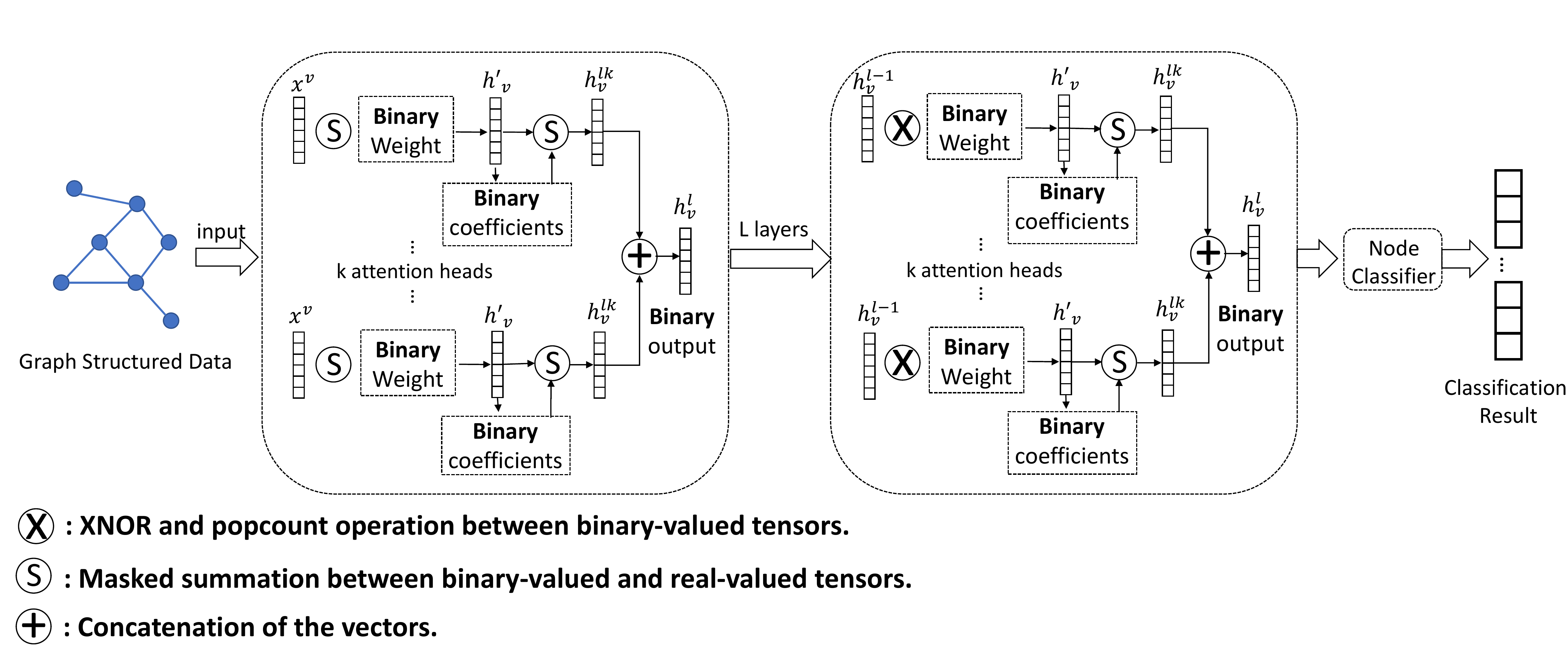}
\vspace{0mm}
\caption{The overall framework of the proposed model BGN. (a) All input node features are projected into a unified representation space by binary-valued weights.(b) Masked summation between binary matrix and real-valued matrix is employed to speed up the dot product. (c) Binary attention coefficients are produced based on the hidden representations. (d) Output of the layer is calculated via multi-head attention mechanism. (e) xnor and popcount are employed to calculate the dot product between binary-valued matrix. (f) Loss calculation and end-to-end optimization for the node classification task.}
\vspace{2mm}
\label{fig:framework}
\end{figure*} 

As illustrated in Figure~\ref{fig:framework},
we introduce a new graph neural network with binarized weights and activations.
Our model BGN ({\bf \uline{B}}inarized {\bf \uline{G}}raph Neural {\bf \uline{N}}etwork) is based on the attention mechanism and can be easily adapted into other graph neural network frameworks.
For a given graph, BGN takes the nodes and their contexts including feature and neighborhood structure information as input. 
Binarization function will transform the weights, activations and even coefficients into binarized vectors to reduce the time and space complexity, while the attention mechanism enables the nodes to attend over their neighborhoods' features.
We also apply the balance function to ensure that $+1$ and $-1$ are almost equal with each other in the binarized vectors.
Furthermore, the gradient estimator is used for backpropagation of gradients through discretization.

The following subsections present the listed key components of our model:
\begin{itemize}
\item Section \ref{sec:framework} introduces the {\bf framework} of our work. 
\item Section~\ref{sec:binarization} introduces the {\bf binarization} of our model in detail, including the {\bf forward propagation} and {\bf backpropagation}.
\item Section \ref{sec:opt} describes the {\bf optimization objective} of our model.
\item Section~\ref{sec:tech} introduces the techniques we used to reduce the time and space complexity and improve the performance.
\item Section~\ref{sec:adapt} introduces the {\bf adaptation} of our model to other GNN frameworks.
\end{itemize}

\subsection{Framework}
\label{sec:framework}
Algorithm \ref{alg:framework} illustrates the framework of our model.
We follow the attention mechanism introduced in \cite{DBLP:conf/nips/VaswaniSPUJGKP17,velivckovic2017graph} to involve the \textit{importance} of the node's neighborhoods into the graph representation learning process.
Given a graph $G(\mathcal{V},E)$, where $\mathcal{V}$ and $E$ denote the set of graph nodes and edges respectively,
we use nodes features $\{ {\bf x}_v, \forall v \in \mathcal{V}\} , {\bf x}_v \in \mathbb{R}^m$ and the neighborhood information of nodes $\{ {\bf \eta}_v, \forall v \in \mathcal{V}\}$ as inputs.
Our model will first produce the binarized node representations ${\bf h}^b_v \in \{+1, -1\}^d$ for each node within the input graph.
After that, the binarized node embeddings will be fed into the output layer to compute the loss for some specific tasks like node classification.

\begin{algorithm}[tb]
\caption{Binarized Graph Neural Network}
\label{alg:framework}
\begin{flushleft}
\textbf{Input}: Graph $G(\mathcal{V},E)$, node features $\{{\bf x}_v, \forall v \in \mathcal{V}\}$, number of layers $L$, binarization function $\mathcal{B}(\cdot)$, number of attention heads $K$, neighbors of node $\{\eta_v, \forall v \in \mathcal{V}\}$\\
\textbf{Output}: Classification result $\{{\bf C}_v, \forall v \in \mathcal{V}\} $\\
\end{flushleft}
\begin{algorithmic}[1] 
\State Let ${\bf h}_v^0 = {\bf x}_v, \forall v \in \mathcal{V}, \forall u \in \eta_v$;
\For{$l = 1,2,\dots , L-1$}
\State $\mathcal{\alpha}^{l}_{uv} = \mathcal{B}(Softmax_{v}({\bf W}^{l}_{0}{\bf x}_u, {\bf W}^{l}_{0}{\bf x}_v))$
\For{$v \in \mathcal{V}$}
\For{$k \in K$}
\State ${\bf h}^{k}_{v} = \mathcal{B}(\sum\limits_{u \in \eta_v}\alpha^{k}_{uv}{\bf W}^{l}_{k}{\bf x}_{u})$;
\EndFor
\State ${\bf h}^{l}_{v} = {Concat}^{K}_{k=1}({\bf h}^{k}_{v})$
\label{alg:agg}
\EndFor
\EndFor
\State $\mathcal{\alpha}^{L}_{uv} = \mathcal{B}(Softmax_{v}({\bf W}^{L}_{0}{\bf h}^{L-1}_u, {\bf W}^{L}_{0}{\bf h}^{L-1}_v))$
\For{$v \in \mathcal{V}$}
\State ${\bf C}_v = Softmax(\sum\limits_{u \in \eta_v}\alpha^{k}_{uv}{\bf W}^{L}{\bf h}^{L-1}_{u})$
\EndFor
\label{alg:get_embedding}
\State \textbf{return} ${\bf C}_v$, the classification result for node $v 
 \in \mathcal{V}$
\end{algorithmic}
\end{algorithm}

\noindent \textbf{Attention Mechanism}
Our proposed framework is based on the graph attention mechanism.
The attention layer is utilized in our model to learn the importance of every node to other nodes.
The key is to get the \textit{importance} of one node's feature to other nodes that is the attention coefficients of the input graph, afterwards, the node's feature can attend on other nodes.
Inspired by~\cite{velivckovic2017graph}, we perform \textit{masked attention} to the model to keep the structural information of the input graph.
Only the attention coefficients of one node with its neighborhood nodes i.e., $\alpha_{ij}, v_j \in \eta_{i}$ will be computed.

In order to obtain the attention coefficients, we use a shared {\bf binarized}  weight matrix ${\bf W} \in \{+1, -1\}^{m \times d'}$ to apply the linear transformation to each node.
Softmax function is used to normalize the coefficients, but unlike the model proposed in~\cite{velivckovic2017graph}, LeakyRelu activation is not employed in our model while the sign function is used to binarize the attention coefficients.
With the following Equation~(\ref{eq:att_coef}), we will get a {\bf binarized} attention coefficient matrix $\mathcal{A} \in \{+1, 0, -1\}^{N \times N}$ where $\mathcal{\alpha}_{ij}$ is the element of the matrix $\mathcal{A}$ (0 is contained in the matrix since we only compute the attention coefficients between neighbors such that the matrix is sparse).
\begin{equation}
\label{eq:att_coef}
\mathcal{\alpha}_{ij} = \mathcal{B}'(Softmax_{j}({\bf W}{\bf x}_i, {\bf W}{\bf x}_j))
\end{equation}
where $\mathcal{B}'$ is the binarization function for attention coefficients which maps 0 to 0, positive values to $+1$ and negative values to $-1$.

Once the attention coefficient matrix is obtained, it will be used to compute the output of the attention layer.
The attention coefficients will multiply the linear transformed node feature.
We employ the \textit{multi-head attention mechanism} to stabilize the learning process.
The binarization function, which is served as activation function, is applied to every attention head to binarize the pre-activations.
And concatenation of the output of K independent attention head is the output of the attention layer.
Therefore, the output node representation will be like following:
\begin{equation}
\label{eq:output}
{\bf h}_{i} = \mathbin{\Vert}^{K}_{k=1}\mathcal{B}(\sum\limits_{j \in \eta_i}\alpha^{k}_{ij}{\bf W}^{k}{\bf x}_{j})
\end{equation}
Where $\mathbin{\Vert}$ means the concatenation of the vectors and ${\bf h}_i$ is the output {\bf binarized} node representation where ${\bf h}_{i} \in \{+1, -1\}^d$.

After several attention layers, the node representation will be fed into the last layer to calculate the loss for specific task which is classification in this paper.
We will introduce the learning objective in the Section~\ref{sec:opt}.

\subsection{Binarization}
\label{sec:binarization}
In this section, we introduce how to obtain a graph neural network with binary parameters that can learn binary representations.
Section~\ref{sec:forward} introduces the binarization function used to  transform the real-valued parameters and pre-activations into binary space.
Section~\ref{sec:backward} introduces the gradient estimators that enable the binarized model to be optimized by the off-the-shelf optimizers such as Adam and SGD.
%
%
\subsubsection{Forward Propagation}
\label{sec:forward}
Binarization function is important in our model.
Specific binarization function will be chosen in the forward propagation calculation process to binarize the weights and the activations.
In that way the low-bit parameters and activations will help to reduce the time and space complexity.
In our case, various binarization functions will work, and the most straightforward example is the sign function.
As mentioned in \cite{courbariaux2016binarized} and \cite{rastegari2016xnor}, deterministic and stochastic binarization based sign function are widely applied to the continuous pre-activations as well as the real-valued weights to obtain binarized activations and weights.
\begin{equation}
\label{eq:det}
\mathcal{B}_{det}(x) = \begin{dcases} +1 & x \geq 0 ,\\
-1 & else,
\end{dcases}
\end{equation}
The above equation is the deterministic binarization function, where $x$ is the real-valued variable.
The stochastic binarization is the sign function with probability:
\begin{equation}
\label{eq:stoch}
\mathcal{B}_{stoch}(x) =
\begin{dcases}
+1 & \text{with probability $p = \sigma (x)$,} \\
-1 & \text{with probability $1 - p$,}
\end{dcases}
\end{equation}
where $\sigma$ denotes the sigmoid function, that is $\sigma(x) = 1/(1+exp(-x))$.
The stochastic binarization is more appealing but needs the computer  to generates random bits while the deterministic binarization is easier to calculate.
Deterministic binarization function(i.e., Equation (\ref{eq:det})) is applied for the binarization of weights and activations because the deterministic sign function provides more stable and reproducible results. 
Please note that we use a variant of deterministic sign function which maps 0 to 0 to binarize the attention coefficients.

\subsubsection{Backprobagation}
\label{sec:backward}
In this part, we describe how to backpropagate the gradients through the binarization function.
We adapt the gradient estimator into our model for better optimization.

\noindent \textbf{Propagation gradients through binarization function.}
It is obvious that the binarization function has zero derivative almost everywhere,
which leads to the zero gradients of the loss function w.r.t the pre-activations and weights.
The trainable variables cannot be updated with zero gradient.
Therefore, the model cannot be trained by simple backpropagation, and the estimation of the gradients should be obtained for optimization.
Previous studies have investigated how to propagate gradients through stochastic discrete functions.
Below we investigate two popular unbiased gradient estimators for binarization function:
straight through estimator and REINFORCE estimator~\cite{DBLP:journals/ml/Williams92}.

\noindent \textbf{Straight through estimator.}
The straight-through estimator is proposed a simple unbiased gradient estimator.
It estimates the derivative of binarization function $\mathcal{B}({\bf h})$ of pre-activation or weight ${\bf h}$ as $\mathbf{1}$ (a vector or matrix whose elements are all 1).
Let ${\bf h}^b$ denote the binarized representation and ${\bf h}$ denote the pre-activation before binarization.
The straight-through estimation of the gradient of the loss $L$ w.r.t the pre-activation ${\bf h}$ is thus:
\begin{equation}
\label{eq:stestimator}
g_h = \frac{\partial L}{\partial {\bf h}} =\frac{\partial L}{\partial \mathcal{B}({\mathbf h})} \cdot \frac{\partial \mathcal{B}({\bf h})}{\partial {\bf h}} = \frac{\partial L}{\partial {\bf h}^b}\mathbf{1}= g_{h^b}\mathbf{1}
\end{equation}
This gradient will then be back-propagated to obtain the gradient of quantities (i.e., pre-activations or weights) that influence ${\bf h}$.

\noindent \textbf{REINFORCE estimator.}
The reinforce estimator is proposed in~\cite{bengio2013estimating} to estimate the expectation of
the gradient $\frac{\partial L}{\partial {\bf h}}$ of loss $L$ with regard to the pre-activation vector or weight ${\bf h}$.
When binarization function $\mathcal{B}(\cdot)$ is stochastic with the probability given by sigmoid, it has been proven that:
\begin{equation}
\label{eq:reinestimator}
\mathbb{E}( \frac{\partial L}{\partial {\mathbf h}}) = \mathbb{E}[(\mathcal{B}({\mathbf h})- \sigma({\bf h}))(L-c)]
\end{equation}
where $\sigma$ is the sigmoid function and $c$ is a constant vector.
To minimize the variance of the estimation, $c$ can be chosen as:

\begin{equation}
\label{eq:c}
c = \frac{\mathbb{E}[(\mathcal{B}({\bf h})- \sigma({\bf h}))^2L]}{\mathbb{E}[(\mathcal{B}({\bf h})- \sigma({\bf h}))^2]}
\end{equation}

The reinforce estimator can work directly on the weights and pre-activations without actual computation of the gradient.
The estimation is obtained by monitoring numerator and denominator during the training process.

Compared with straight through estimator, reinforce estimator is more advanced with better performance in many applications. However, we observe that its performance is not superior than the straight through estimator. 
On the other hand, straight through estimator helps the model to obtain the gradient faster than the reinforce estimator due to its simplicity.
The comparison between these two gradient estimators with regards to the performance is included in Section~\ref{sec:experiment}.
In practice, we choose straight through estimator for our model in the experiments.


\subsection{Optimization Objectives}
\label{sec:opt}
Existing GNN-based graph embedding approaches provide an end-to-end model, which focuses on the node classification task.
Therefore, our model is also learned for the node classification task.
Below, we introduce the objective of BGN and the learning process that optimizes the parameters.

For the node classification learning, we feed the binarized embedding ${\bf h}^b_v$ into the output layer to predict the class label for the node.
The predicting probability of label ${\bf C}_i$ is written as:
\begin{equation}
\label{eq:softmaxprob}
p({\bf C}_{vk} \mid {\bf h}^b_v) = Softmax^{k}_{\zeta}(\sum\limits_{u \in \eta_v}\alpha^{L}_{uv}{\bf W}^{L}{\bf h}^{b}_{u})
\end{equation}
where $\zeta$ denotes the number of labels for each node.
After obtaining the classification result in Equation~(\ref{eq:softmaxprob}), we calculate the cross-entropy as the loss for the node classification task.

\begin{equation}
\label{eq:lossclass}
L_{class} =-\sum_{v \in \mathcal{V}_{labeled}}\sum_{k = 1}^\zeta {\bf C}^{\mathcal{L}}_{vk}\log({\bf C}_{vk})
\end{equation}
where $\mathcal{V}_{labeled}$ is the set of nodes that have label information which are used for training process, ${\bf C}^{\mathcal{L}}_{vk}$ is the multi-hot encoding for ground truth classification labels.

The gradients will be back propagated via estimator and be applied on the optimization of parameters by the off-the-shelf optimizer during the training process.

\subsection{Techniques to Improve the Model}
\label{sec:tech}

\begin{figure}[tb]
\vspace{-2mm}
\centering
\includegraphics[width=0.95\columnwidth]{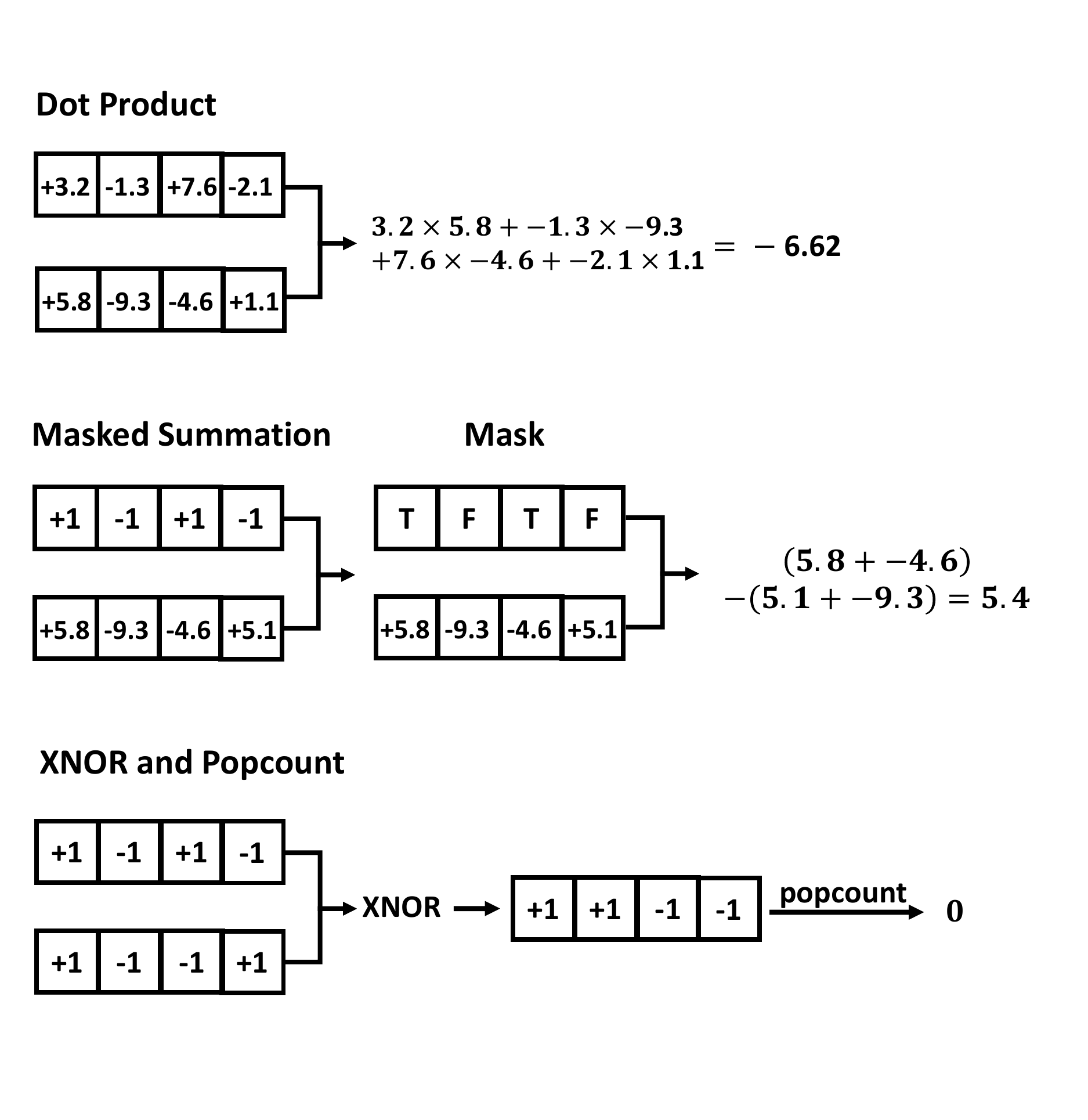}
\vspace{-10mm}
\caption{The toy examples of (a) dot product (b) Masked summation and (c) \textit{xnor} and \textit{popcount} instruction}
\label{fig:dot}
\vspace{-2mm}
\end{figure}

Several techniques are used on binarized graph neural network model to reduce the time and space complexity and improve the performance. 
Logic operation \textit{xnor} between binary values, build-in CPU instruction \textit{popcount} and the masked summation are used to replace the tradition arithmetic operation dot product to reduce time complexity. 
The figure~\ref{fig:dot} is a toy example that introduces the differences between these operations.
Balance function is used to make $+1$ and $-1$ to be balanced in the embedding vectors which can raise the performance of the GMN.
Also, the binary parameters of the neural network and the binary node representations can reduce the space complexity intuitively.

\subsubsection{xnor and popcount}
\label{sec:xnor}
The logic \textit{xnor} and CPU build-in instruction \textit{popcount} between binary matrices are used to replace the dot product between them.

\begin{table}[tb]
\centering
\begin{tabular}{cc|c}
\hline
{\bf Input A}  & {\bf Input B} & {\bf Output}\\
\hline
{\bf +1}  &{\bf +1} & {\bf +1}   \\
{\bf +1} &{\bf -1} & {\bf -1} \\
{\bf -1} & {\bf +1} & {\bf -1} \\
{\bf -1} & {\bf -1} & {\bf +1} \\
\hline
\end{tabular}
\caption{\textit{xnor} calculation}
\label{tab:xnor}
\vspace{-5mm}
\end{table}

As shown in Table~\ref{tab:xnor}, \textit{xnor} produces binary value with input of $+1$ and $-1$. 
Instruction \textit{popcount} is then be employed to count the number of bits that is set to $1$.
The \textit{xnor} can be more than one order of magnitude faster than the dot product which can dramastically reduce the time complexity.
As mentioned in \cite{courbariaux2016binarized}, a 32-bit floating point multiplier costs about 200 Xilinx FPGA slices, whereas a 1-bit \textit{xnor} gate costs only 1 slice.

\subsubsection{Masked Summation}

Masked summation is used to replace the dot product between binary  matrix and real-valued matrix.
The binary matrix will be transformed into the mask matrix with "True" and "False".
During the multiplication, the real-valued vector will be masked by the corresponding mask vector, then the positive and negative masked vector are produced with only the elements at the same position as "True" and "False" on the mask vector.
The model calculates the summations of the positive and negative masked vector separately.
The subtraction of these two summation results is the result of dot product between the given matrices.

The masked summation can reduce the time complexity of dot product of two matrix.
Usually, the time complexity of naive dot product between two real-value matrices $M_1 \in \mathbb{R}^{m\times n}$ and $M_2 \in \mathbb{R}^{n\times d}$ is $\mathcal{O}(mnd)$,
while the time complexity of \textit{masked summation} between binary matrix $M_1 \in \{-1, +1\}^{m\times n}$ and real-valued matrix $M_2 \in \mathbb{R}^{n\times d}$ is $\mathcal{O}(nd)$.
Theoretically and also in practice, the masked summation can significantly reduce the time complexity of our proposed binarized graph neural network.

\subsubsection{Balance Function}
The distribution of $+1$ and $-1$ is sometimes unbalanced in the representation vectors.
For example, if most pre-activations ${\mathbf h}$ have \textit{positive} elements, the output graph representation vector of binarization function ${\mathbf h}^b$ will be formed mainly by $+1$. Then the dot product of two vectors will be $d$ which is the dimension of the vectors.
This unwanted situation should be avoid because it dramatically lower the effectiveness of the proposed model, especially when the BGN is applied to GMN which requires a great number of dot product between representations.
As a result, we apply the following balance function to the pre-activations before binarization in order to balance the distribution of \textit{positive} and \textit{negative} elements of pre-activations:

\begin{equation}
\label{eq:balance}
\textit{Balance}({\bf h}) = {\bf h} - \overline{\bf h}
\end{equation}

Where the $\overline{\bf h}$ is the vector whose elements are all mean value of the pre-activation vector ${\bf h}$.
The balance function ensures that the pre-activation vectors contain almost half \textit{positive} and half \textit{negative} elements, which leads to the balance distribution of $+1$ and $-1$ after binarization.

\subsection{Adapted to Other GNN Based Models}
\label{sec:adapt}
The proposed binarized graph neural network is a very general framework that can be adapted to other graph neural network-based model to project the real-valued parameters and activations into the  binary space to reduce the space and time cost.
We introduce how we binarize the state-of-the-art GNN-based model AS-GCN~\cite{DBLP:conf/nips/Huang0RH18} and the graph matching network.


\subsubsection{Binarization of AS-GCN}
\noindent AS-GCN is a general framework that is designed for fast representation learning based on graph neural networks such as GCN.
Therefore, the binarization of AS-GCN is similar to our proposed BGN.
We use deterministic binary function to binarize the parameters and pre-activations of AS-GCN.
And straight through estimator is employed for back propagation.
The binarized model is denoted as BGN-ASGCN in our experiment.

\subsubsection{Binarization of GMN}
\noindent As mentioned above, the time cost of GMN comes mainly from the pair-wise node similarity computation. 
We utilize the deterministic binarization function (Equation~(\ref{eq:det})) on the preactivations and transform the node and graph representations into binary codes such that the xnor can be applied to replace the dot product.
Straight through estimator (Equation~(\ref{eq:stestimator})) is used for the back propagation.
Furthermore, we noticed that the distribution of $+1$ and $-1$ is usually not symmetric which dramatically lower the performance, hence, balance function (Equation~(\ref{eq:balance}))is employed on the graph representations. 

%

\section{Experiment}
\label{sec:experiment}

We conduct extensive experiments to evaluate the performance of our model for the node classification task on real-world network datasets.
We compare the time and space efficiency thoroughly between the proposed model and other baseline models.
The case study shows the effectiveness and efficiency brought by our framework on the GNN-based application such as GMN.

\subsection{Dataset}
\label{sec:dataset}
To facilitate the comparison between our model and the relevant baselines, we conduct the classification experiments on three well-known citation network datasets:
{\bf Cora}, {\bf Citeseer} and {\bf Pubmed} \cite{sen2008collective}.
Each dataset contains bag-of-words representations of documents and citation links between the documents.
Graph $G$ is constructed based on the citation links.
In the classification task, we only use $20$ labeled instances per class for training.
The test data contains 1000 nodes as in GCN, GAT and AS-GCN.

The details of the datasets are summarized in the Table\ref{tab:datasets}.

\begin{table}[tb]
\centering
\begin{tabular}{cccccc}
\hline
{\bf Dataset}  & {\bf \#Nodes} & {\bf \#Edges} & {\bf \#Classes} & {\bf \#Labled Nodes} \\
\hline
Cora       & 2708  & 5429 & 7   & 140     \\
Citeseer & 3327 & 4732 & 6  & 120\\
Pubmed       & 19717  &44338 & 3 & 60      \\
\hline
\end{tabular}
\caption{Citation Datasets}
\label{tab:datasets}
\vspace{-6mm}
\end{table}

\subsection{Baseline Methods}
\label{sec:baseline}
\vspace{1mm}
The following GNN-based and binary embedding methods are compared as baselines: 

\noindent \textbf{GCN} (Graph Convolutional Network)~\cite{velivckovic2017graph} is a semi-supervised neural network method for node classification.

\noindent \textbf{GAT} (Graph Attention Network)~\cite{velivckovic2017graph} is a graph neural network model which first exploits the attention mechanism to solve the node classification task.

\noindent \textbf{AS-GCN} (Adaptive Sampling over GCN)~\cite{DBLP:conf/nips/Huang0RH18} is a state-of-the-art method for node classification task. AS-GCN aims to increase the scalability of GCN using adaptive sampling.
The experiments demonstrate that the application of BGN can further reduce the time and space complexity of AS-GCN.

\noindent \textbf{GAT-binary} and \textbf{ASGCN-binary} are the models that directly apply sign function on the node representations learned by the original version of GAT and AS-GCN. The naively binarized representations will be fed into the task-specific layer to learn the classification result.

\noindent \textbf{GAT-tanh} and \textbf{ASGCN-tanh} are the models that employ the binarization function \textit{tanh} used by \textit{DeepMind}'s work. \textit{tanh} function is used to binarize the parameters and embedding vectors of GAT and AS-GCN. We clip the value of the parameters and activations in both models to make sure that \textit{tanh} can produce “exact” binary codes.

\noindent \textbf{INH-MF}~\cite{DBLP:conf/kdd/Lian0ZGCT018} is a MF-based information network hashing algorithm that learns binary codes as node embedding which can preserve high-order proximity.

\noindent \textbf{BANE}(Binarized Attributed Network Embedding)~\cite{DBLP:conf/icdm/YangPLZ18} is an extension of DNE~\cite{DBLP:conf/ijcai/PLOS18} which based on the Weisfeiler-Lehman proximity matrix factorization learning function to produce binary node representations.

\subsection{Experiment Setup}
\label{sec:setup}

For the performance experiment, we evaluate the models with the same bit-width representations.
For the experiment of inference efficiency, the embedding dimension of our method and other baseline methods are all set to 64. 
During training process, the whole graph can be seen, but only a few nodes are labeled while most nodes have no label information.
We put all nodes information in one training phase due to the need of calculation for graph attention coefficients.

For this classification task, we report the average accuracy of the evaluated GNN-based embedding approaches after ten independent runs using the accuracy metric introduced in \cite{kipf2017semi,velivckovic2017graph}. 
Because INH-MF and BANE only produce the binary embedding vectors but have no build-in classifier,  we employ the one-vs-rest logic regression implemented by Liblinear ~\cite{DBLP:journals/jmlr/FanCHWL08} to obtain the classification result of the networks, in which 90\% nodes are labeled.

All the experiments were conducted on the server which is running RHEL 7.5 and has 2x 2.4GHz Intel Xeon  E5-2680 v4 (14 Cores) CPU, 256GB 2400MHz ECC DDR4-RAM and 2x NIVDIA Quadro P5000 16GB Graphics Card (GPUs) (2560 Cores).

\subsection{Classification Results}
\label{sec:classification}

Because our model produces the compact representations for vertices, we compare the performance between our model and other baselines with the same bit width.




\subsubsection{Comparison Among Binary Embedding Methods}
\label{sec:class_binary}
We compare the classification results between our model and other binary-valued embedding methods.

\begin{figure}[tb]
\vspace{-2mm}
\centering
\includegraphics[width=0.9\columnwidth]{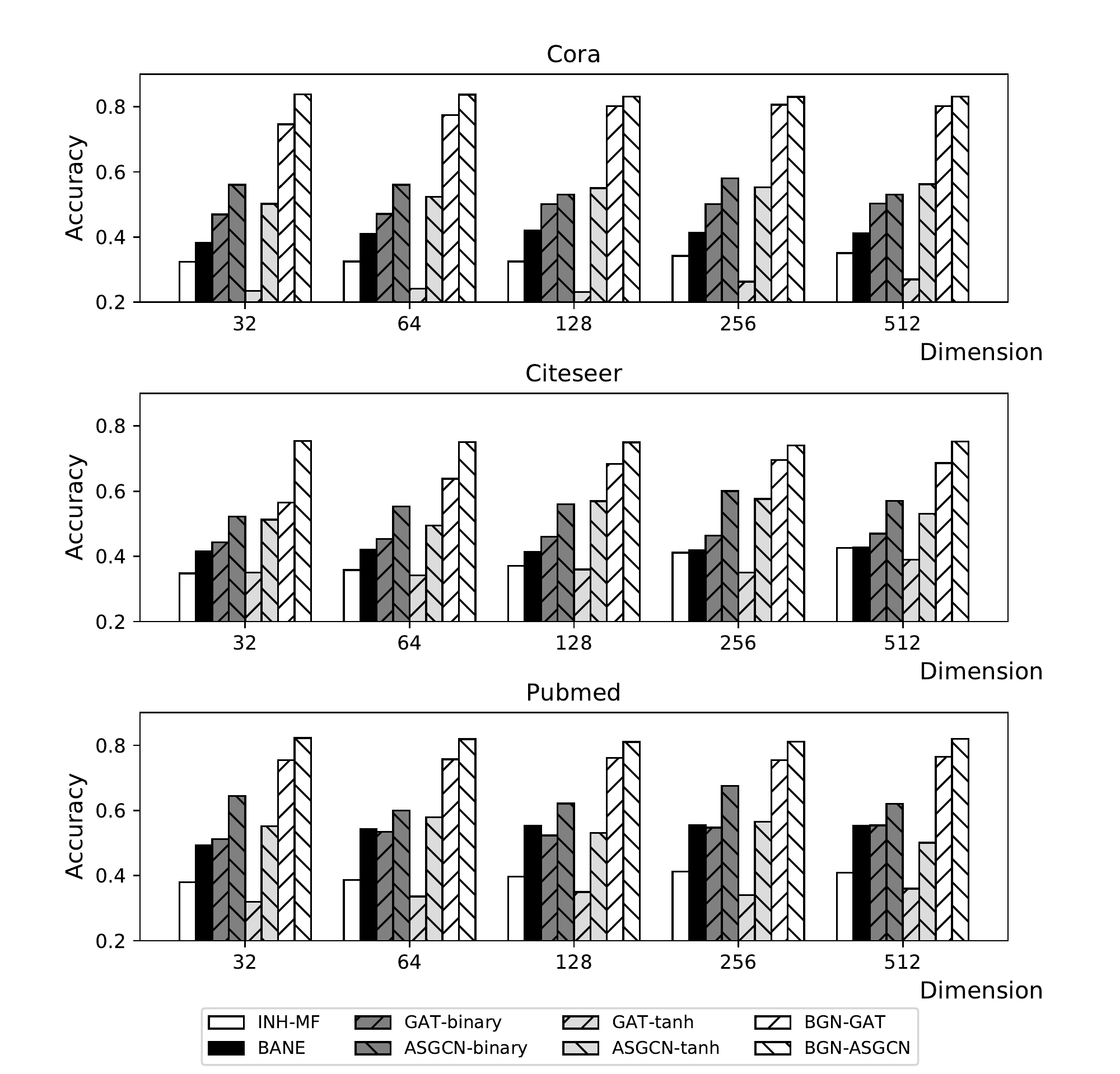}
\vspace{-4mm}
\caption{Classification results of three citation network dataset among the binary-valued embedding methods with different embedding dimensions}
\vspace{-6mm}
\label{fig:binary_class}
\end{figure} 

As shown in the Figure~\ref{fig:binary_class}, under different embedding dimensions, BGN outperforms all the other binary-valued embedding methods significantly on all three datasets.
With the help of the graph neural network, our model can make better use of the graph structured data and feature information and is trained specifically for the node classification task.
Therefore, our model outperforms other MF-based binarized graph embedding models by a significantly large margin.
In comparison with the naively binarized GAT-binary and ASGCN-binary, our model considers the binary property of parameters and vectors during the training process, hence our model achieves better accuracy.
In terms of GAT-tanh and ASGCN-tanh, because \textit{tanh} function has zero gradient when the output is nearly $+1$ or $-1$ and has real-value output when the gradient is not zero.
This property determines that \textit{tanh} function is not suitable for binarize the neural network.
When the input values are clipped to produce exact binary parameters and embeddings via \textit{tanh} function, the gradient will be zero which results in the insufficient optimization and worse performance than BGN.

\subsubsection{Comparison among the GNN-based methods}
\label{sec:GNN}
\begin{figure}[tb]
\vspace{-2mm}
\centering
\includegraphics[width=\columnwidth]{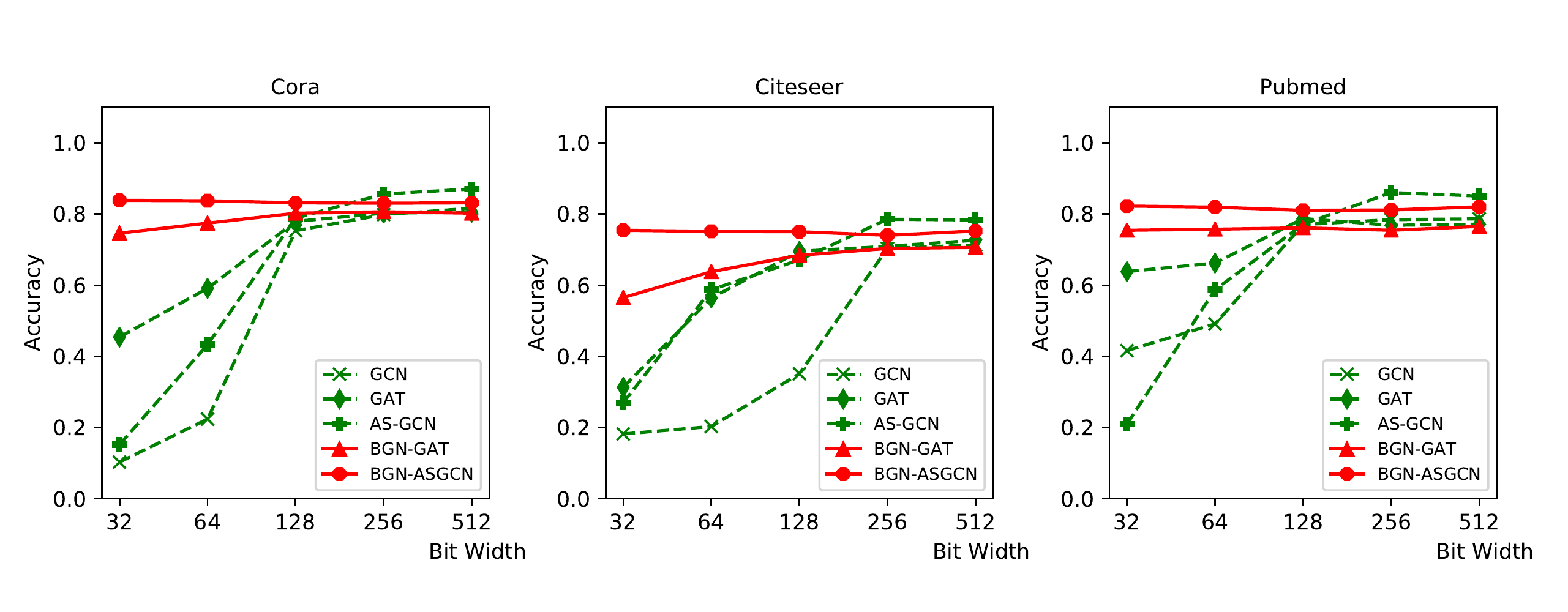}
\vspace{-6mm}
\caption{Classification results of three citation network dataset among the GNN-based methods with varied bit width for embedding vector}
\vspace{-4mm}
\label{fig:com_gnn}
\end{figure}

We compare our model with other GNN-based methods (GCN, GAT and AS-GCN). 
All baseline methods produce the real-valued embedding vectors each dimension of which is encoded by at least 32 bits.
Compared with these methods, each dimension of the embedding vectors learned by our model is only encoded by 1 bit.
As a result, a real-valued 16 dimension vector requires at least 256 bits while a binary vector only requires 16 bits.
Figure~\ref{fig:com_gnn} shows the performance of the models with bit width varies for a single embedding vector.

Our model significantly outperforms all the baseline methods with low bit width.
When getting more space for the learned representations, our model can still achieve competitive classification results compared with the state-of-the-art graph neural network-based methods.
In conclusion, the performance gap between our model and baselines with large bit-width representations is acceptably small while our model's performance is notably better with the low bit-width representations.

\subsection{Comparison of Time and Space Efficiency}
\label{sec:time}

\begin{table*}[tb]
\centering
\begin{tabular}{m{0.09\columnwidth}<{\centering}|p{0.12\columnwidth}<{\centering}|p{0.12\columnwidth}<{\centering}|p{0.12\columnwidth}<{\centering}|p{0.12\columnwidth}<{\centering}|p{0.12\columnwidth}<{\centering}}
\toprule[1pt]
\multicolumn{2}{c|}{\bf Dataset} & {\bf GAT} & {\bf AS-GCN} &{\bf BGN-GAT} &{\bf BGN-ASGCN}\\
\midrule
\multirow{3}*{Cora}& Time(s) & $1.9 \times 10^{-1}$ & $1.0\times 10^{-1}$ & $1.0 \times 10^{-2}$& $8.0\times 10^-3$\\
~& Space(bit) & $2.46\times 10^8$ & $3.04\times 10^6$& $1.32\times 10^7$ & $1.97\times 10^5$\\
~&Accuracy & 84.0\% & 87.3\%& 77.7\% & 84.1\%\\
\hline
\multirow{3}*{Citeseer} & Time(s) & $2.8 \times 10^{-1}$ & $2.8\times 10^{-1}$ & $1.4\times 10^{-2}$ & $1.8 \times 10^{-2}$\\
~ & Space(bit) & $7.60\times 10^6$ & $7.83\times 10^6$ & $2.49\times 10^5$ & $4.86\times 10^5$ \\
~ & Accuracy & 72.1\% & 78.9\% & 63.7\% & 77.2\%\\
\hline
\multirow{3}*{Pubmed} & Time(s) & $3.8 \times 10^{1}$ & $4.54\times 10^{0}$ & $1.1 \times 10^{0}$ & $2.1\times 10^{-1}$\\
~ & Space(bit) & $1.03\times 10^6$ & $1.06\times 10^6$ & $3.85\times 10^4$ & $7.01\times 10^4$ \\
~ & Accuracy & 78.2\% & 89.0\% & 75.7\% & 82.0\%\\
\bottomrule[1pt]
\end{tabular}
\caption{Comparison of performance, inference time and memory space required for the parameters between the real-valued and BGN-based models.}
\label{tab:space}
\vspace{-4mm}
\end{table*}

In this section, we report the inference time and space efficiency of our model.
The inference is the process that produces the classification result when we have already trained the model.
Acceleration is brought by the \textit{xnor} and \textit{popcount} operation with just little sacrifice on the classification performance.
In this experiment, we train the binary parameters and activations of our model, then replace dot product operation between binarized matrices by \textit{xnor} and \textit{popcount} and also replace the dot product between binary matrix and real-valued matrix by masked-summation during the inference process.

Table~\ref{tab:space} reports the experiment results. 
Our model under the binarized framework is more than one order of magnitude faster than the baseline methods GAT and AS-GCN with regards to the inference time.
The proposed model can be up to $29\times$ faster and save up to $28\times$ space compared with the baseline methods.


\subsection{Analysis of Binarization}
\label{sec:analysis}

In this section, we introduce the effect of the estimator and binarization level with regard to the space, time and performance.
We compare the space, inference time and performance between BGN-GAT and GAT on the Cora dataset.
We fix the dimension of embedding vector to 64 for both methods and change the setting of BGN to show the space and time saving compared with the baseline GAT.

\begin{table}[tb]
\centering
\resizebox{\columnwidth}{20mm}{
\begin{tabular}{c|ccccc}
\toprule[1pt]
{\bf Method}  & {\bf Estimator} &{\bf Param space}& {\bf Vec space} &{\bf Speed up}& {\bf Accuracy}\\
\midrule
GAT & N/A& $1\times$&$1\times$& $1\times$&84.0\%\\
${\rm BGN}^{w}$&STE&$1/28$& $1\times$&$3.7\times$&80.5\%\\
${\rm BGN}^{w}$&Reinforce&$1/28$& $1\times$&$3.8\times$&80.3\%\\
${\rm BGN}^{e}$&STE&$1\times$& $1/1.02$&$1.3\times$&81.2\%\\
${\rm BGN}^{e}$&Reinforce&$1\times$& $1/1.02$&$1.2\times$&81.3\%\\
${\rm BGN}^{we}$&STE&$1/28$& $1/1.02$&$5.7\times$&77.2\%\\
${\rm BGN}^{we}$&Reinforce&$1/28$& $1/1.02$&$6.1\times$&77.5\%\\
${\rm BGN}^{wec}$&STE&$1/28$& $1/19$&$18.7\times$&77.7\%\\
${\rm BGN}^{wec}$&Reinforce&$1/28$& $1/19$&$19.1\times$&76.9\%\\
\bottomrule[1pt]
\end{tabular}}
\caption{Trade-off between time/space efficiency and classification accuracy of proposed BGN w.r.t the level and setting of binarization. }
\vspace{-4mm}
\label{tab:analysis}
\end{table}

Result is shown in Table~\ref{tab:analysis} where ${\rm BGN}^{w}$, ${\rm BGN}^{e}$, ${\rm BGN}^{we}$ and ${\rm BGN}^{wec}$ mean that the BGN is with weights binarized, embedding vectors binarized, weights and embedding vectors binarized, weights, embedding vectors and attention coefficients binarized based on the graph attention mechanism respectively.
We can conclude from the Table~\ref{tab:analysis} that 
(1) when the weights, activations and attention coefficients are all binarized, the BGN-GAT can save largest space for parameters and the output vectors while holding acceptable classification accuracy.
(2) Straight through estimator and reinforce estimator have similar accuracy on the node classification task. Therefore, we choose the STE for our model in the above experiments because of its simplicity and certainty.
(3) Compared with original GAT, BGN-GAT can save $28 \times$ space for model parameters, $19 \times$ space for activations and achieve $19 \times$ speed up.

\subsection{Case Study}
\label{sec:case}
In this section, we investigate how binarized graph neural network improve the time efficiency of the GNN-based applications such as GMN.
Because GMN needs to compute the pair-wise dot product between node and graph embedding vectors, the time consumption is extremely high when the number of nodes in each graph goes up.
However, with the binary representations, we can apply \textit{xnor} between binary vectors to replace the dot product, which will alleviate the time complexity problem significantly.
The following experiment results will introduce the performance and time complexity of GMN with binary node and graph representations compared with the origin version.
The graph similarity will then be used for the graph matching task.

\vspace{1mm}
\noindent \textbf{Experiment Setup}

\noindent We follow the experiment setting of \cite{DBLP:conf/icml/LiGDVK19} to test the performance of Binarized GMN.
The training data is generated by sampling binomial graphs $G_1$ with $n$ nodes and edge probability $p$~\cite{erdHos1960evolution}.
Then the positive example $G_2$ is generated by randomly substituting $k_p$ edges from $G_1$ with new edges and negative example $G_3$ is generated by substituting $k_n$ edges from $G_1$, where $k_p < k_n$.
In the experiment, we set $k_p = 1$, $k_n = 2$ and $p = 0.2$.
We also set the hamming similarity between vectors as loss function, which is more suitable for the binary-valued vectors as the loss function to train the model.
The model needs to predict a higher similarity score for positive pair $(G_{1}, G_{2})$ than negative pair $(G_{1}, G_{3})$.
The evaluation metric remains the same: (1) pair AUC - the area under the ROC curve for classifying pairs of graphs as similar or not on a fixed set of 1000 pairs and (2) triplet accuracy - the accuracy of correctly assigning higher similarity to the positive pair in a triplet than the negative pair on a fixed set of 1000 triplets.

\vspace{1mm}
\noindent \textbf{Inference time and Graph Matching Performance}

\noindent We report the graph matching accuracy and inference time of the binarized and original GMN with regards to the number of nodes in each graph.
The default setting in GMN is 20 nodes per graph, which is quite small for real-world networks.
We set the number of nodes in one graph from 20 to 160 and keep other settings the same as described above to evaluate the performance and inference time.
The dimensions of node and graph representations are set to 32 and 64 respectively.

\begin{figure*}[tb]
\vspace{-2mm}
\centering
\includegraphics[width=0.95\columnwidth]{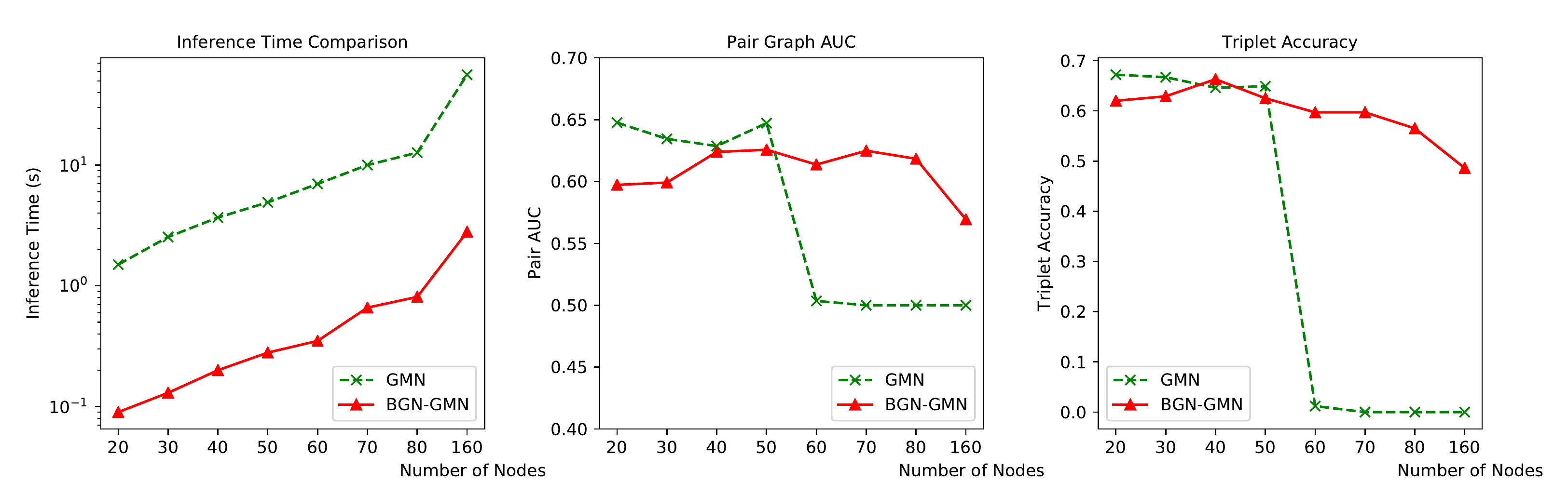}
\vspace{-2mm}
\caption{The performance of graph matching and inference time for GMN and BGN-GMN w.r.t the number of nodes per graph}
\vspace{-2mm}
\label{fig:gmn_runtime}
\end{figure*} 

As shown in Figure~\ref{fig:gmn_runtime}, the inference of BGN-GMN is significantly faster. 
This is because of the fact that the similarity computation (pair-wise dot product) between node representations of two graphs mainly accounts for the time complexity of GMN.
Under the same dimension of node and graph embedding vectors, BGN-GMN is up to $21\times$ faster than the baseline model in terms of the inference time with the help of the replacement of dot product by fast operations such as \textit{xnor} and \textit{popcount} between binary vectors.

In terms of graph matching task, the original version of GMN has better performance when the number of nodes in each graph is small.
However, when the number of nodes gets larger, the pair AUC and triplet accuracy will both decay. 
When the number of nodes is more than $60$, the real-valued representations cannot tell the similarity difference between the graphs. 
Hence, the model is not able to learn the different similarity scores for positive and negative pairs of graphs with the hamming similarity metric.
However, with the help of binarization and balance function, the binary representations still hold an acceptable and more robust performance for the graph matching task.
This is due to the fact that the binarized model produces \textit{true} binary representations for the calculation of hamming loss and is designed for the graph matching task specifically on hamming space.

\vspace{1mm}
\noindent \textbf{Parameter Sensitivity Analysis}

\noindent We compare the performance of binarized and original version GMN to show the effect of dimension for node and graph embedding vectors.
We set the number of nodes in each graph $n=30$ for this comparison.
We change the dimension of graph embeddings produced by two models to ensure them to produce the same bit-width embedding vectors and keep the other settings as the same to compare the performance of two models.

\begin{figure}[tb]
\centering
\subfigure[Performance w.r.t bit width of {\bf Graph} Representation]{
\begin{minipage}[htb]{0.95\columnwidth}
\centering
\includegraphics[width=\columnwidth]{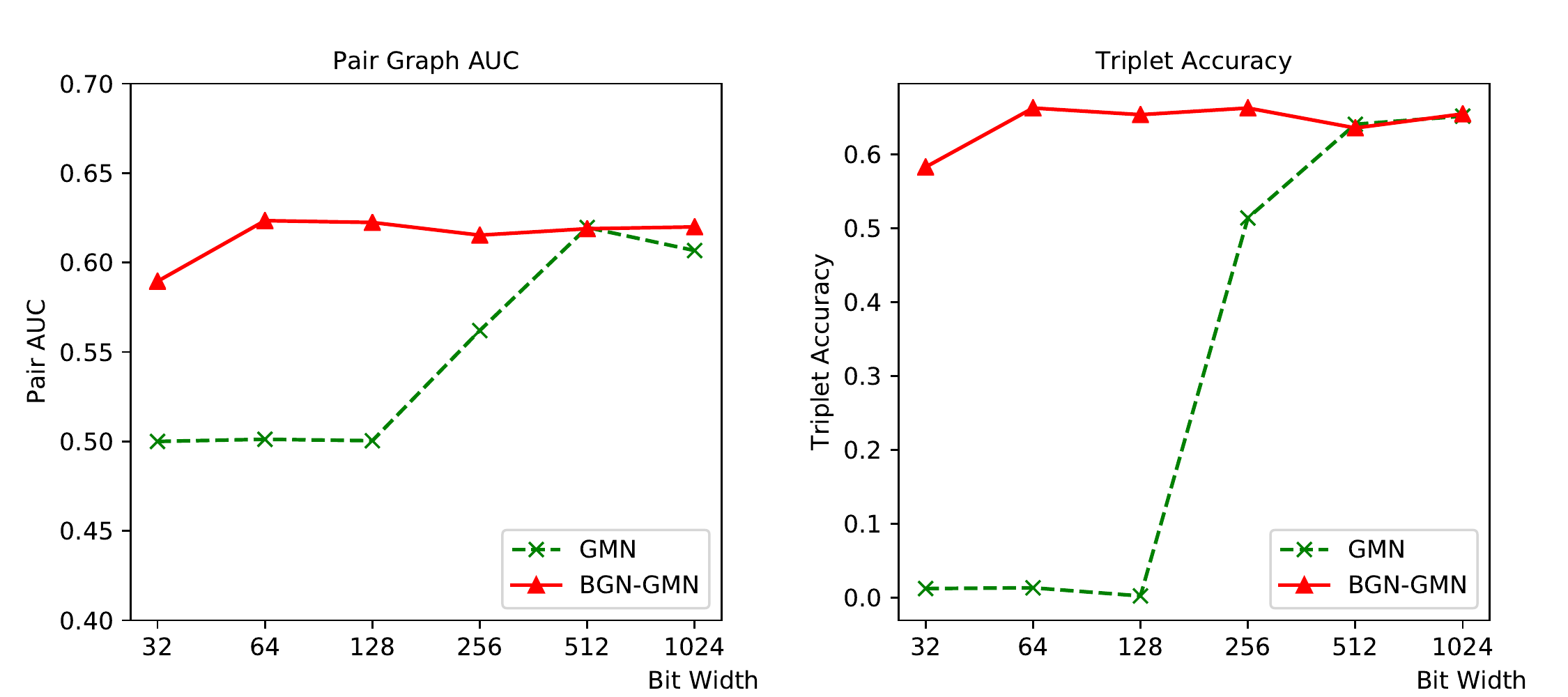}
\end{minipage}
\label{fig:gmn_bit}
}
\subfigure[Performance w.r.t bit width of {\bf Node} Representation]{
\centering
\begin{minipage}[htb]{0.95\columnwidth}
\includegraphics[width=\columnwidth]{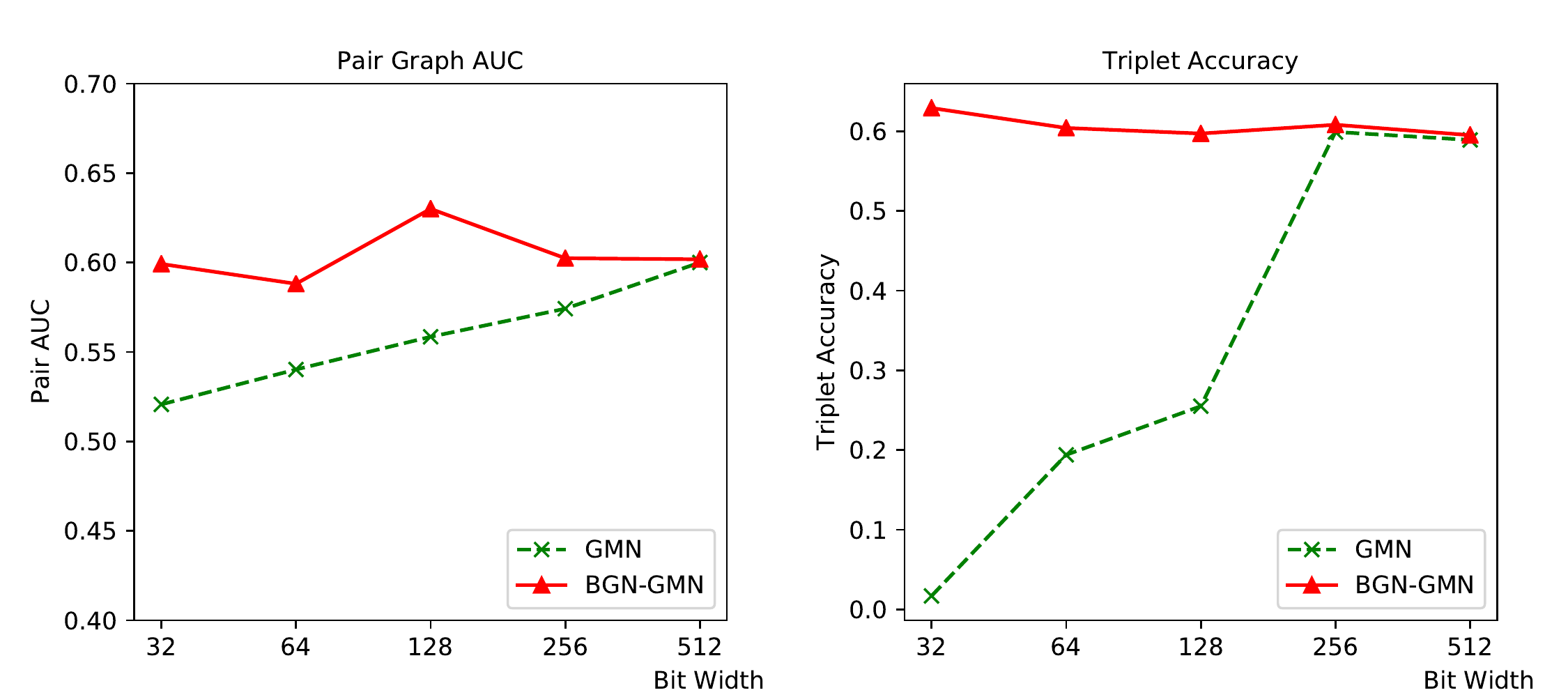}
\end{minipage}
\label{fig:gmn_node}
}
\vspace{-2mm}
\caption{The performance comparison of graph matching task between original version of GMN and the BGN-GMN with (a) graph representations binarized and (b) node representations binarized}
\vspace{-6mm}
\end{figure}

The result is included in Figure~\ref{fig:gmn_bit}. We can find that the binary graph representations tend to have better performance when they are low bit-width and have similar accuracy when the bit-width for the representations getting larger.
The binary representations have more robust performance compared with the baseline model when the dimension of embedding varied.

The node representations' binarization is more important than the graph representations' because the dot product operation is mainly conducted between the node representations which costs plenty of time.
The performances of GMN and BGN-GMN are compared under different bit-width for the node embedding vectors by varying the dimensions.

As shown in Figure~\ref{fig:gmn_node}, the result for the pair-wise AUC is similar between the binary and the real-valued node embedding vectors, but BGN-GMN holds a better performance with low bit-width representations.
As for the triplet graph accuracy, the binary embedding vector achieves better performance with short code length and similar accuracy as real-valued node embedding with long code length.
These results indicate that the binary representations are much better for the comparison between two graphs under low bit-width circumstances.
In line with the result of the binary graph embedding vectors, the binary node embedding vectors also have more robust performance compared with the real-valued node representations.

%
%

\section{Conclusion}
\label{sec:conclusion}

We present a model focused on the challenging problem of seeking binary representations of network embeddings using a compact neural network structure.
We proposed a novel binarized graph embedding method, namely BGN, that has binarized parameters and enables GNNs to learn discrete embedding.
The binarized neural network can reduce the memory and time cost of the GNN such that increases the scalability of GNNs.
BGN can be naturally integrated into other GNN models to enhance the performance of the model such as graph matching network in terms of the inference time and space consumption.
External experiment also illustrates that BGN can increase the time efficiency while holding competitive accuracy.

%
%

\bibliographystyle{spmpsci}      
\bibliography{reference}   

%
%

\end{document}